\documentclass{article}

\usepackage[preprint]{neurips_2025}

\usepackage[utf8]{inputenc} % allow utf-8 input
\usepackage[T1]{fontenc}    % use 8-bit T1 fonts
\usepackage{hyperref}       % hyperlinks
\usepackage{url}            % simple URL typesetting
\usepackage{booktabs}       % professional-quality tables
\usepackage{amsfonts}       % blackboard math symbols
\usepackage{nicefrac}       % compact symbols for 1/2, etc.
\usepackage{microtype}      % microtypography
\usepackage{xcolor}         % colors

\usepackage{array}
\usepackage{graphicx}
\usepackage{amsmath}
\usepackage{amsthm}
\usepackage{amssymb}
\usepackage{tikz}
\usetikzlibrary{automata, arrows, positioning}
\usepackage{subfigure}
\usepackage{algorithmic}
\usepackage{algorithm}
\usepackage{amsmath,amsfonts,bm}

\def\eqref#1{equation~\ref{#1}}
\def\1{\bm{1}}

\DeclareMathAlphabet{\mathsfit}{\encodingdefault}{\sfdefault}{m}{sl}
\SetMathAlphabet{\mathsfit}{bold}{\encodingdefault}{\sfdefault}{bx}{n}

\newcommand{\R}{\mathbb{R}}

\DeclareMathOperator*{\argmax}{arg\,max}

\usepackage{amsmath}
\usepackage{amssymb}
\usepackage{mathtools}
\usepackage{amsthm}
\newtheorem{theorem}{Theorem}
\newtheorem{corollary}{Corollary}
\newtheorem{lemma}{Lemma}
\newtheorem*{theorem*}{Theorem}
\newtheorem*{corollary*}{Corollary}
\newtheorem*{lemma*}{Lemma}

\newtheorem{proposition}{Proposition}

\newtheorem*{property*}{Property}

\newtheorem{example}{Example}

\newcommand{\ie}{\text{i.e.}, }

\DeclarePairedDelimiter\norm{\lVert}{\rVert}% || ||
\DeclarePairedDelimiter\innorm{\langle}{\rangle}% < >

\DeclareMathOperator{\St}{\mathcal{S}}
\DeclareMathOperator{\A}{\mathcal{A}}

\title{Policy Gradient with Tree Search:\\ Avoiding Local Optimas through Lookahead}

\author{
Uri Koren\thanks{Equal contribution} \\
Technion\\
\texttt{uri.koren@campus.technion.ac.il} \\
\And
Navdeep Kumar* \\
Technion \\
\texttt{navdeepkumar@campus.technion.ac.il} \\
\AND
Uri Gadot \\
Technion \\
\texttt{uri.gad@campus.technion.ac.il} \\
\And
Giorgia Ramponi \\
University of Zurich \\
\texttt{giorgia.ramponi@uzh.ch}\\
\AND
Kfir Levy \\
Technion \\
\texttt{kfirylevy@technion.ac.il}  \\
\And Shie Mannor \\
Technion \\
\texttt{shiemannor@gmail.com} \\
}

\begin{document}

\maketitle

\begin{abstract}
 Classical policy gradient (PG) methods in reinforcement learning frequently converge to suboptimal local optima, a challenge exacerbated in large or complex environments. This work investigates Policy Gradient with Tree Search (PGTS), an approach that integrates an $m$-step lookahead mechanism to enhance policy optimization. We provide theoretical analysis demonstrating that increasing the tree search depth 
$m$-monotonically reduces the set of undesirable stationary points and, consequently, improves the worst-case performance of any resulting stationary policy. Critically, our analysis accommodates practical scenarios where policy updates are restricted to states visited by the current policy, rather than requiring updates across the entire state space. Empirical evaluations on diverse MDP structures, including Ladder, Tightrope, and Gridworld environments, illustrate PGTS's ability to exhibit "farsightedness," navigate challenging reward landscapes, escape local traps where standard PG fails, and achieve superior solutions.
\end{abstract}

\section{Introduction}
% RL and PG in general
Reinforcement Learning (RL) has become a cornerstone in artificial intelligence, enabling agents to solve sequential decision-making problems in a wide range of domains, from robotics to healthcare to autonomous driving \cite{Sutton1998}. Among the plethora of RL algorithms, policy gradient (PG) methods stand out for their ability to directly optimize policies and scale to high-dimensional action spaces \cite{sutton1999policy,schulman2015trust}. Their success in landmark applications such as AlphaGo, robotics, and game-playing agents underscores their versatility and potential \cite{AlphaZero,mnih2015humanlevel}.

Despite these successes, PG methods are not without challenges. A fundamental limitation arises from the non-concave nature of the objective in policy space \cite{valuePolytope}, making PG methods susceptible to convergence at suboptimal local minima \cite{PGTS, PG_local_conv_single_time_scale,actor-critic}. As global convergence PG requires the strict assumption of finite mis-match coefficient or initial state-distribution have strictly positive probability at all states.\cite{PG_ConvRate, SoftMaxPG_conv,PG_Conv_JLiu_Elementaryanalysispolicygradient_linearConv}.  This issue is exacerbated in real-world applications where large and continuous state spaces, combined with safety constraints, restrict exploratory behavior. For example, in autonomous driving, policies cannot feasibly explore all potential states, nor is it desirable to employ aggressive exploratory strategies that might compromise safety. These practical constraints render the standard assumptions for global convergence of PG methods—such as uniform exploration or a bounded mismatch coefficient—highly intractable \cite{SoftMaxPG_conv,PG_conv_average_reward}.

Popular variants of policy gradient (PG) algorithms, such as TRPO and PPO \cite{trpo}, employ trust region updates to improve stability. While these methods enhance convergence robustness, they remain susceptible to local optima. Exploratory techniques like $\epsilon$-greedy PG and entropy-regularized methods can potentially improve solution quality by promoting exploration. However, these undirected exploration methods often require exponential time to sufficiently explore high-diameter Markov Decision Processes (MDPs) \cite{PureExploration}.

Another line of work incorporates Monte Carlo Tree Search (MCTS) to unroll the running policy across multiple steps, improving the estimation of Q-values and leading to better performance \cite{AlphaZero}. While MCTS reduces the variance of updates and facilitates faster, more stable convergence, it cannot escape the inherent limitation of converging to suboptimal stationary policies characteristic of standard PG methods.

Lookahead extensions of policy iteration and policy mirror descent \cite{Efroni2018BeyondTO,protopapas2024policymirrordescentlookahead} have demonstrated faster convergence to global optima compared to non-lookahead variants. However, these methods require policy updates at all states during each iteration—a condition that is computationally infeasible for large-scale MDPs. In real-world scenarios, the state space may be prohibitively large or only partially known, making full-state updates impractical both in terms of exploration and memory requirements.
The Policy Gradient with Tree Search (PGTS) method, as proposed in \cite{PGTS}, demonstrates convergence to a globally optimal policy without requiring any conditions on the mismatch coefficient. While the original formulation relies on infinite tree search depth—a strong theoretical result—it is computationally impractical for real-world applications. Our work explores the practical utility of PGTS with finite tree depths across diverse MDP structures, including random MDPs, Ladder MDPs, Tightrope MDPs, and Gridworld MDPs. We observe that while PGTS behaves similarly to standard PG in well-connected random MDPs, it significantly outperforms PG in large-diameter MDPs such as Ladder and Gridworld. Notably, PGTS not only identifies higher-quality solutions but also accelerates the discovery of high-reward regions, leading to faster convergence.

A key advantage of PGTS is highlighted in Tightrope MDPs, where its far-sighted strategy enables exploration of regions with temporary declines in returns, subsequently recovering and converging to optimal policies faster than standard PG. This contrasts with the monotonic improvement seen in PG, which may be shortsighted in such settings.

While PGTS may not guarantee global optimality under local updates, we theoretically demonstrate that the number of stationary points decreases with increased tree depth. This property implies a reduction in local optima, enhancing solution quality as the tree depth grows. To our knowledge, this is the first bottom-up approach that systematically addresses the limitations of local optima in PG methods, contrasting with top-down approaches that impose restrictive global convergence assumptions \cite{SoftMaxPG_conv,PG_conv_average_reward,PG_ConvRate,Natural_Actor_Critic_Conv_Bhatnagar}.

We believe that the development of sample-based PGTS methods compatible with deep neural networks has the potential to significantly enhance the performance of reinforcement learning in practice. However, understanding the convergence rate of PGTS remains a challenging task. We hypothesize that these rates may depend on structural properties of the MDP, such as its hardness, diameter, or connectivity—a direction we leave open for future exploration.

\section{Preliminaries}
% \subsection{Markov Decision Process}
A Markov Decision Process (MDP) is defined as a tuple $(\mathcal{S}, \mathcal{A}, P, R, \gamma, \mu)$, where the components are as follows: $\mathcal{S}$ and $\mathcal{A}$ represent the state and action spaces; $P \in (\Delta_{\mathcal{S}})^{\mathcal{S} \times \mathcal{A}}$ is the transition kernel; $R \in \mathbb{R}^{\mathcal{S} \times \mathcal{A}}$ is the reward function; $\gamma \in [0, 1)$ is the discount factor; $\mu \in \Delta_{\mathcal{S}}$ is the initial state distribution; and $\Delta_{\mathcal{X}}$ denotes the probability simplex over the set $\mathcal{X}$ \cite{puterman2014markov, Sutton1998}. The set of policies $\Pi$ includes all policies $\pi \in \Pi$, where $\pi(a|s)$ is the probability of taking action $a$ in state $s$. We use the shorthand notations $P^\pi(s'|s) = \sum_{a} \pi(a|s) P(s'|s, a)$ and $R^\pi(s) = \sum_{a} \pi(a|s) R(s, a)$ to represent the state transition probabilities and the expected reward under policy $\pi$, respectively. The objective in an MDP is to find a policy $\pi \in \Pi$ that maximizes the return \[J^\pi := \mu^T (I - \gamma P^\pi)^{-1} R^\pi = E\Bigm[\sum_{k=0}^{\infty}\gamma^kR(s_k,a_k)\mid s_0\sim \mu,\pi,P\Bigm].\] 
For shorthand $J^*=\max_{\pi}J^\pi$ denotes the global optimal return and $\pi^*$ is global optimal policy that achieves this optimal return.
The first-order (one-step) gradient of the return is given by:
\begin{align}
    \frac{\partial  J^\pi}{\partial \pi(a|s)} = d^\pi(s) Q^\pi(s, a)
\end{align}
where $d^\pi = \mu^T (I - \gamma P^\pi)^{-1} = E[\sum_{k=0}^\infty\gamma^k\mathbf{1}(s_k)\mid s_0\sim \mu,\pi,P]$ is the occupancy measure where $\mathbf{1}(s)\in\R^{\St}$ is one-hot indicator vector for coordinate $s$, further, $Q^\pi = R + \gamma P v^\pi$ is the action-value (Q-value) function, and $v^\pi = (I - \gamma P^\pi)^{-1} R^\pi$ is the state-value function \cite{Sutton1998}. The update rule for policy gradient is then:
\begin{align}\label{eq:PG}
   \pi_{k+1} = \text{proj}_{\Pi}\left(\pi_k +\eta_k\frac{\partial J^{\pi_k}}{\partial \pi}\right), 
\end{align}
where $\eta_k$ is learning rate and $proj_{\Pi}$ is projection operator on the policy space $\Pi$. This rule has been demonstrated to converge to a globally optimal solution under the condition of a bounded mismatch coefficient, 
$\max_{\pi}\norm{\frac{d^{\pi^*}}{d^\pi}},$
which is guaranteed if $
\Tilde{\mu} := \min_{s} \mu(s) > 0 $
\cite{agarwal2021theory, xiao2022convergence, PG_Conv_JLiu_Elementaryanalysispolicygradient_linearConv}.
Notably, \(\Tilde{\mu} > 0\) also ensures 
$
\min_s d^\pi(s) > 0
$
allowing for policy updates across all states at each iteration, effectively making the policy gradient (PG) method resemble a soft policy iteration

However, satisfying the \(\Tilde{\mu} > 0\) condition is challenging in practice, especially in large Markov Decision Processes (MDPs), especially with continuous state spaces. Moreover, even if we employ exploratory policies such as $\epsilon$-greedy (taking random action with probability $\epsilon$) the mismatch coefficient could be exponential in state space for high-diameter MDPs such as Ladder MDPs \cite{PGTS}. And without this assumption, the policy update rule can become trapped in suboptimal local maxima, as illustrated by the Ladder MDP in Example 1 in \cite{PGTS}.
% \textbf{PGTS.} To mitigate this issue, \cite{kumar2024policy} proposed a tree search inspired $m-$step look-ahead policy gradient update rule:
% \begin{align}\label{PGTS}
%     \pi_{k+1} = proj\Bigm[\pi_k +\eta_k U^m_{\pi_k}\Bigm],
% \end{align}
% where the PGTS (policy gradient with tree search) $U^m_{\pi}$ is defined as 
% \begin{align*}
%   U^m_\pi(s,a) &= d^{\pi}(s)\max_{\pi_1,\cdots,\pi_m}\E\Bigm[ \sum_{n=0}^{m-1} \gamma^{n}R^{\pi_n}(s_n)  \\&\qquad+ \gamma^{m}\sum_{a_m}\pi_m(s_m,a_m)Q^{\pi}(s_m,a_m)\mid s_0=s,a_0=a,P\Bigm]\\ &= d^\pi(s)\Bigm(T^m Q^\pi\Bigm) (s,a),\qquad \text{(from Lemma 1 of \cite{PGTS})}
% \end{align*}
% where $T$ is the optimal bellman operator defined as $(TQ)(s,a) = R(s,a) + \gamma \sum_{s'}P(s'|s,a)\max_{a'} Q(s',a')$.

% % \textbf{Explaination/Intuition  of the update PGTS}

% \begin{proposition}[Theorem 1 of \cite{PGTS}] For infinite depth ($m\to\infty$) and infinite learning rate ($\eta_k \to \infty$), the PGTS update rule \eqref{PGTS} converges to global optimal policy $\pi^*$ in $O(S)$ iterations, that is,
% \[ \pi_{m} = \pi^*, \qquad \forall m\geq S.\]
% \end{proposition}

\section{Method}
In this work, we investigate the Policy Gradient with Tree Search (PGTS) algorithm (see Algorithm \ref{main:alg:PGTS}), originally introduced by \cite{PGTS}, and explore its theoretical properties and insightful mechanisms. The PGTS algorithm incorporates an $m$-step look-ahead (tree search) and is governed by the following update rule: 
\begin{align}\label{PGTS}
    \pi_{k+1}(\cdot|s) = \text{proj}\Big[\pi_k(\cdot|s) + \eta_k d^{\pi_k}(s) \big(T^mQ^{\pi_k}\big)(s,\cdot)\Big],
\end{align}  
where \(\eta_k\) is the learning rate, \(\text{proj}\) denotes the projection operator onto the probability simplex, and \(T\) is the Bellman operator for Q-values, defined as  
$(TQ)(s,a) := R(s,a) + \gamma \sum_{s'} P(s'|s,a) \max_{a'} Q(s',a').$  For \(m=0\), the PGTS update rule reduces to the standard policy gradient method \cite{PolicyGradient}, as \(T^0\) is conventionally the identity operator \(I\). Furthermore, the PGTS update direction can be rewritten in an expanded form that highlights its connection to tree search:  
\[
d^\pi(s)(T^mQ^\pi)(s,a) = d^\pi(s) \max_{\pi_1,\ldots,\pi_m} E\Big[\sum_{k=0}^{m-1} \gamma^k R(s_k,a_k) + \gamma^m Q^\pi(s_m,a_m)\Big],
\]  
where \(s_0 = s\), \(a_0 = a\), \(a_k \sim \pi_k(\cdot|s_k)\), and \(s_{k+1} \sim P(\cdot|s_k,a_k)\) for all \(k\), as shown in Lemma 1 of \cite{PGTS}. While the tree search involves maximization over policies \(\pi_1, \ldots, \pi_m\) and may seem computationally expensive, its complexity is efficiently managed. Using the Bellman operator, the computation of \(T^m Q\) scales linearly with \(m\), making the process tractable even for deeper look-aheads.

Now, we present the Ladder MDP (Example 1 from \cite{PGTS}), where the standard policy gradient (PG) fails significantly, and none of its variants, including TRPO, PPO, or MCTS, can escape the local optima. Further, exploratory strategies such as epsilon-greedy policy can take exponential (in state space) time to find the reward state.

\begin{example}[Example 1 of \cite{PGTS}]\label{ex:ladder}
The ladder MDP as illustrated below has fives states and are two actions, 'left' and 'right' in each state. The reward is  zero except unit reward at  state $s_4$ under the action 'right'. The initial policy $\pi_0$ be to always play the action 'left'.

\begin{center}
\begin{tikzpicture} [node distance = 2.4cm, on grid, auto]
\centering
\node (s0) [state, initial text = {}] {$s_0$};
\node (s1) [state, right = of s0] {$s_1$};
\node (s2) [state, right = of s1] {$s_2$};
\node (s3) [state, right = of s2] {$s_3$};
\node (s4) [state, right = of s3] {$s_4$};
 
\path [-stealth, thick]
    (s0) edge [loop left]  node {left}()
    (s0) edge [bend left] node  {right} (s1)
    (s1) edge [bend left] node {left}   (s0)
    (s1) edge [bend left] node  {right} (s2)
    (s2) edge [bend left] node {left}   (s1)
    (s2) edge [bend left] node  {right} (s3)
    (s3) edge [bend left] node {left}   (s2)
    (s3) edge [bend left] node  {right} (s4)
    (s4) edge [bend left] node {left}   (s3)
    (s4) edge [loop right]  node {right,+1}();
\end{tikzpicture}

\end{center}
\end{example}

\begin{figure}[h]
    \centering
    \includegraphics[width=\linewidth]{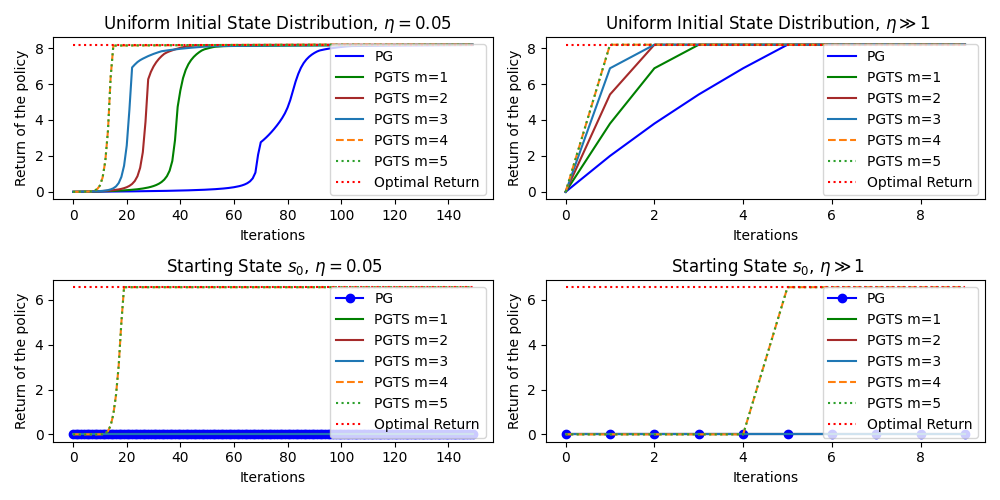}
    \caption{PGTS on Ladder MDP: For initial state $s_0$, PG and PGTS are stuck at local solution until depth $3$.}
    \label{fig:ladder}
\end{figure}
For the initial state $s_0$ (where $\mu(s_0) = 1$), it is evident that the initial policy $\pi_0$ is a local maximum, satisfying $\frac{\partial J^{\pi_0}}{\partial \pi} = 0$. This indicates that the standard policy gradient will get stuck at $\pi_0$, which is a local maximum, while the globally optimal policy always plays the action 'right'. 

The reason the policy gradient is zero at $\pi_0$ lies in its one-step look-ahead nature. The one-step gradient search fails to reveal better policies from $\pi_0$ because these policies are not visible within a single step. Consequently, the policy gradient converges to the local solution, yielding a return of zero. In contrast, PGTS requires a look-ahead depth of at least $m=4$ to achieve global convergence, as illustrated in Figure \ref{fig:ladder}.

Furthermore, Figure \ref{fig:ladder} demonstrates that a uniform initial state distribution enables PG to find the optimal solution, but its initial progress is notably slow. In contrast, PGTS with higher depth significantly outperforms PG with a uniform state distribution. As illustrated in the figure, greater depth allows PGTS to propagate the high-reward information from state $s_4$ more quickly, resulting in a faster improvement of the policy.
\begin{figure}
    \centering
    \includegraphics[width=\linewidth]{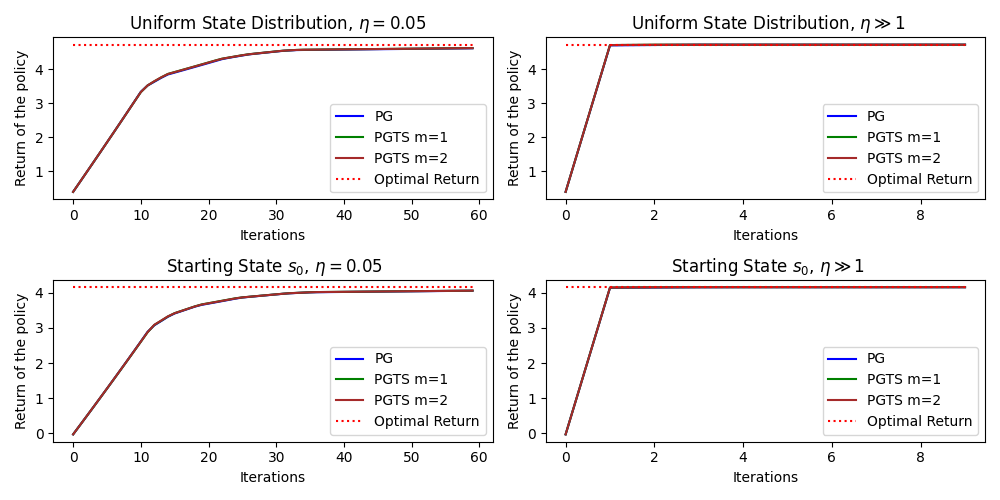}
    \caption{Random MDP: $S=10, A=2,\gamma=0.9$, randomly generated transition kernel and reward. All methods (PG and PGTS) have similar performance as the MDP is  well connected and dense reward.}
    \label{fig:random}
\end{figure}
Interestingly, higher depth than $m=4$ doesn't help further neither with uniform initial state distribution nor with $s_0$ as starting state, intuitively this is probably due to the fact that the diameter of this Ladder  MDP is four. Now, we showcase that the PG performs reasonably well in the random MDPs which are very well connected.   
\begin{example}[Random MDP] This MDP has ten states with two actions each, with randomly generated transition kernel and reward function.
\end{example}
In this random MDP, every state is reachable from every other state under any action, and the rewards are densely distributed. As shown in Figure \ref{fig:random}, PG performs almost as well as PGTS with any depth. This is likely due to the efficient dissemination of information, facilitated by the well-connected structure and dense rewards of these MDPs, in contrast to the Ladder MDPs.

Policy Gradient (PG) is inherently a local method that seeks to greedily improve the policy at each iteration. Specifically, the PG update rule \eqref{eq:PG} guarantees monotone improvement:$$
J^{\pi_{k+1}} \geq J^{\pi_k}
$$for all $k \geq 0$ and any learning rates $\eta_k \geq 0$ \cite{PG_Conv_JLiu_Elementaryanalysispolicygradient_linearConv}. 
Counterintuitively , this property is not preserved for PGTS higher depth $m\geq 1$. Now, we introduce the Tightrope MDP \cite{Efroni2018BeyondTO}, which demonstrates the "farsightedness" of PGTS—an intriguing and distinguishing feature of the algorithm.
\begin{example}[Tightrope MDP \cite{Efroni2018BeyondTO}]  The Tightrope MDP, as illustrated below, consists of four states, each with two available actions. Among these states, two are terminal: $s_2$, which yields a unit reward ($+1$), and $s_3$, which incurs a large negative reward ($-10$). The initial policy prescribes taking the action "left" in all states, whereas the optimal policy involves taking the action "right" in every state.
 
\begin{center}
\begin{tikzpicture}[node distance=2cm, auto, ->, thick]

% Define states
\node[state] (s0) {$s_0$};
\node[state] (s1) [right of=s0] {$s_1$};
\node[state] (s2) [right of=s1] {$s_2$};
\node[state] (s3) [above of=s1] {$s_3$};

% Transitions for "right" (forward direction)
\path (s0) edge[] node {right} (s1);
\path (s1) edge[] node {right} (s2);
\path (s2) edge[loop right] node {left,right,+1} (s2);

% Transitions for "left" (backward direction)
\path (s1) edge[] node {left} (s3);
\path (s3) edge[loop right] node {left,right,-10} (s3);
\path (s0) edge[loop left] node {left} (s0);

\end{tikzpicture}
    
\end{center}
\end{example}
\begin{figure}[h]
    \centering
    \includegraphics[width=\linewidth]{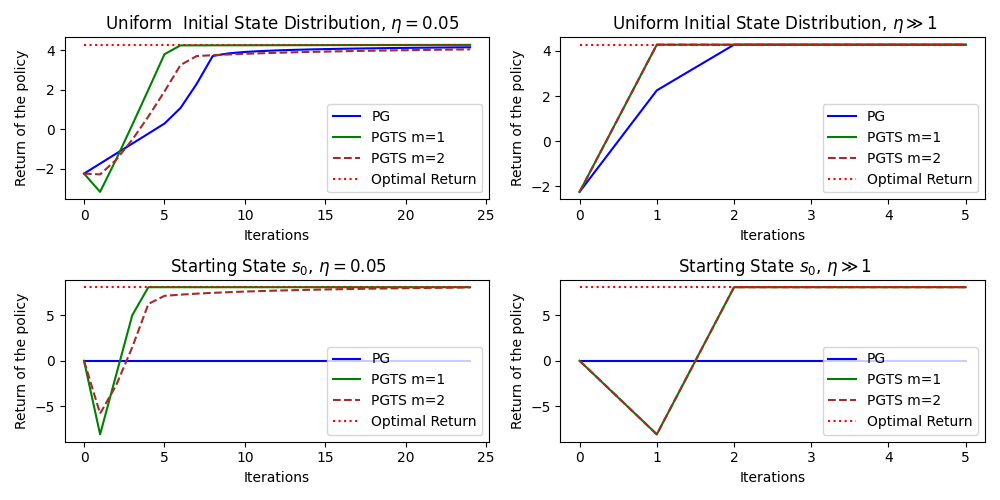}
    \caption{PGTS on Tightrope MDP }
    \label{fig:Tightrope}
\end{figure}
As before, when starting from state $s_0$, PG is unable to discover the positive reward at $s_2$. Consequently, it does not improve and continues taking the action "left" at all states. In contrast, PGTS successfully discovers the optimal policy.

An interesting behavior arises with a uniform initial state distribution. In this case, PG initially takes the action "left" at $s_0$ until the agent updates its policy to take the action "right" at $s_1$ with sufficient probability instead of "left." This behavior makes sense because, at the start, $s_1$ leads to the negative reward state $s_3$, and the small learning rate delays the policy update to favor "right" at $s_1$. Thus, avoiding $s_1$ from $s_0$ in the meantime becomes a reasonable choice for PG, as it greedily improves the performance, as shown in Figure \ref{fig:Tightrope}.

On the other hand, PGTS identifies the positive reward state $s_2$ from the very first iteration and begins updating its policy at $s_0$ to favor the action "right." While this initially decreases the performance of the policy for a few iterations (since the agent frequently visits the negative reward state $s_3$ from $s_1$), PGTS ultimately starts favoring $s_2$ over $s_3$ as it updates the policy at $s_1$. Once this occurs, the return increases rapidly as the agent reaches $s_2$ more efficiently from $s_0$. This behavior reflects PGTS's farsighted nature, as it foresees this scenario and updates the policy accordingly to achieve the optimal goal.

This is evident from the learning curves illustrated in Figure \ref{fig:Tightrope}. PG's return increases monotonically with each iteration due to its greedy updates, while PGTS's return initially decreases but recovers quickly, achieving the optimal return much earlier than PG.

\begin{figure}
    \centering
    \includegraphics[width=\linewidth]{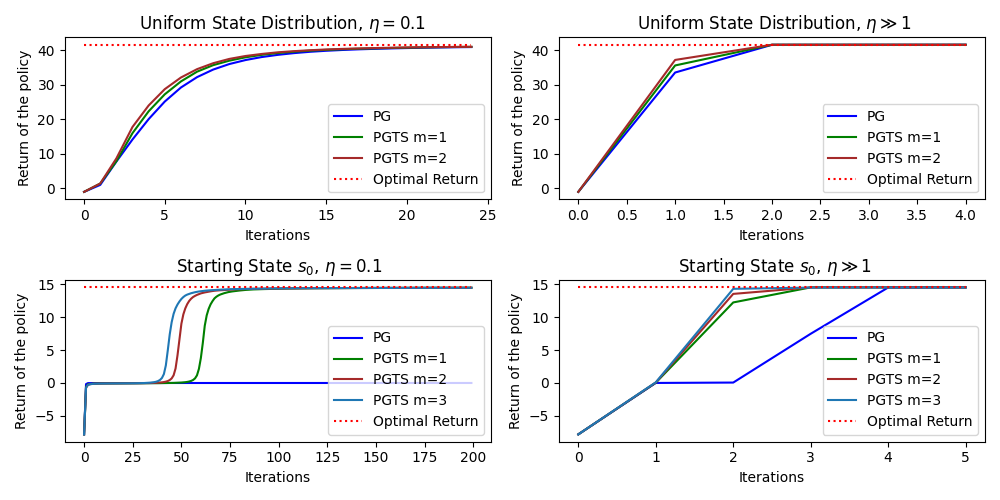}
    \caption{\textbf{PGTS on Grid MDP}}
    \label{fig:grid}
\end{figure}

\begin{example}
The Grid MDP, as described in Table \ref{tb:gridMDP}, is a $10 \times 10$ grid with states labeled from $s_{00}$ to $s_{99}$. It offers four actions: 'left,' 'right,' 'up,' and 'down,' each resulting in intuitive one-step transitions with a probability of $0.99$. With a probability of $0.01$, however, any action causes a random one-step transition. The MDP features a goal state, $s_{99}$, which provides a reward of $+10$, as well as trap states, $s_{12}$ and $s_{32}$, which incur a penalty of $-10$. The initial policy takes all actions uniformly at all states.

\begin{table}[H]
    \centering
\begin{tabular}{|p{0.7cm}|p{0.7cm}|p{0.7cm}|p{0.7cm}|p{0.7cm}|p{0.7cm}|p{0.7cm}|p{0.7cm}|p{0.7cm}|p{0.7cm}|}
\hline $s_{00}$ &  &  &  &  &  &  &  &  & $s_{09}$ \\ \hline
 & Trap  &  &  &  &  &  &  &  &  \\ \hline
 &  &  &  &  &  &  &  &  &  \\ \hline
 &  & Trap  &  &  &  &  &  &  &  \\ \hline
 &  &  &  &  &  &  &  &  &  \\ \hline
 &  &  &  &  &  &  &  &  &  \\ \hline
 &  &  &  &  &  &  &  &  &  \\ \hline
 &  &  &  &  &  &  &  &  &  \\ \hline
 &  &  &  &  &  &  &  &  &  \\ \hline
 $s_{90}$&  &  &  &  &  &  &  &  & Goal \\ \hline
\end{tabular}
\caption{Grid MDP}
    \label{tb:gridMDP}
\end{table}
\end{example}
As illustrated in Figure \ref{fig:grid}, PGTS and PG perform similarly under a uniform initial state distribution on the Grid MDP. However, when starting from $s_{00}$ with a small learning rate of $0.1$, PG prioritizes avoiding nearby traps, which improves immediate returns. In doing so, it becomes overly conservative and shortsighted, ultimately staying at $s_{00}$.

In contrast, PGTS with depth $m \geq 1$ exhibits a more robust learning process. Initially, it avoids the immediate traps and then quickly identifies the goal state $s_{99}$, discovering an optimal path to reach it. This demonstrates the farsightedness of PGTS, enabling it to escape local optima and efficiently achieve the global solution.

We conclude that PGTS demonstrates foresightedness and discovers better policies than PG. Moreover, the hyperparameter depth in PGTS can be increased when tackling difficult problems to find better optima. All codes and results are available at \url{https://anonymous.4open.science/r/pgts-C77F/}. In the next section, we formally establish that number of local optimas that can trap PGTS, decreases with the depth.

\section{ Stationary Points of PGTS}
In this section, first we show that set of stationary points PGTS doesn't depend  on the learning rate but only of depth. Then we establish that the number of possible stationary points of PGTS decreases with the depth. And for infinite depth only stationary policies of PGTS are global optimal policies. Thus implying that the quality of solutions of PGTS improves with depth. Furher, we characterize the nature of stationary policies for PG and PGTS which can inspire in designing better strategies that can avoid local optimas.

\paragraph{Stationary Points of PGTS Don't Depend on Learning Rates.}
  Let us define $\Pi^m_\eta$ to be the set  of all possible  saddle (fixed/stationary) points of the PGTS \eqref{PGTS}, that is  
\begin{align} \label{SaddleSet}
  \Pi^m_\eta = \Bigm\{ \pi\mid \pi = proj\bigm[\pi +\eta
     D^\pi T^mQ^{\pi} \bigm]\Bigm\},
\end{align}
where $D^\pi \in \R^{(\St\times\A)\times (\St\times\A)}$ is diagonal matrix defined as $D^\pi((s,a),(s,a))=d^\pi(s)$. Let  $\sigma_\pi :=\Bigm\{s\in\St\mid d^\pi(s)>0\Bigm\}$ denote the support set of occupation measure of the policy $\pi$. Then the result below states that the all the stationary policies in $\Pi_\eta^m$ has the same properties as the stationary policies in $\Pi^m_\eta$.
\begin{lemma} [Equivalence w.r.t. step size] \label{rs:PiEquivalance} For every $\eta \geq 0$ holds $\pi\in\Pi^m_\eta$ if and only if
\begin{align*}
\pi_s  
\in\argmax_{\pi'_s}\innorm{\pi'_s,(T^mQ^\pi)(s,\cdot)}, \qquad \forall s\in\sigma_\pi
\end{align*}

This implies that the set stationary points of the PGTS \eqref{PGTS} are independent of the learning rates. That is,
    \[\Pi^{m}_\infty = \Pi^{m}_\eta\,\qquad\forall\eta>0.\]

\end{lemma}
Note that different step size schedules $\eta_k$ may converge to different stationary policies, as expected from policy gradient methods in non-convex functions. However, the set of such solutions only depend on the depth of the PGTS. Henceforth,  we drop the sub-script $\eta$, and simply denote $\Pi^m $ as the set of stationary points of the update rule \eqref{SaddleSet}. 

Theorem 1 of \cite{PGTS} established that the set of stationary points for infinite depth $\Pi^\infty $ contains only the global optimal policies $\pi^*$. The result below states that  the global optimal policy $\pi^*$ lies in each set $\Pi^m$ for all depths.

\begin{proposition}\label{prop:allDepthContainOptimal}[All depth can possibly lead to global solutions] For all depth $m\geq 0$, we have
    \[\pi^*\in\Pi^m \]
\end{proposition}

Now the question arises, what are other stationary points? As \cite{PGTS} showed that standard PG (PGTS with depth $m=0$) has lots local stationary points, and infinite depth PGTS has none. Although, it hypothesized that the set of stationary points are monotonically decreasing with depth but the proof was left as a open question. Before we prove this hypothesis, first we discuss the optimality condition of the stationary points of PG and PGTS.

\paragraph{Lookahead leads to more robust stationary points.} It is well known \cite{Sutton1998} that a global optimal policy $\pi^*$ satisfies the following optimality condition:
\[\pi \in \argmax_{\pi'}\innorm{\pi,Q^\pi}\qquad\text{equivalently}\qquad v^\pi(s) = \max_{a}Q^\pi(s,a).\] The result below extends this optimality conditions for the stationary points of PGTS including the standard PG.
\begin{lemma}[PGTS Optimality Condition]\label{rs:optimalityCondn} 
    For $\pi\in\Pi^m$ if and only if
\[v^\pi(s) = \max_{a}(T^mQ^\pi)(s,a),\qquad\forall s\in\sigma_\pi,\]
or equivalently \[\pi \in \argmax_{\pi'}\innorm{\pi',D^\pi T^mQ^\pi},\] where  $D^\pi \in \R^{(\St\times\A)\times (\St\times\A)}$ is diagonal matrix defined as $D^\pi((s,a),(s,a))=d^\pi(s)$.
\end{lemma}
For the special case of $m=0$ (standard policy gradient), the stationary policy $\pi\in\Pi^0$ satisfies
\[v^\pi(s) = \max_{a}Q^\pi(s,a)\]
for only those states $s$ that the policy visits. This is an interesting observation in itself. This can lead to the better development of the exploration strategies, that incentivizes those exploratory actions/strategies that encourages the agent to look outside its horizon. In other words, $\epsilon$-greedy exploration strategy plays a random action with probability $\epsilon$ at each iteration, treating all the actions the same. While the result above suggests that the if those exploratory actions are taken that has transition probability of states that are not visited so far, then this exploration strategy has better chances of escaping local solutions. However, this budding inspiration requires a careful investigation for the fruitful design of better exploratory strategies.

Using this result, we can now show our main theorem - we prove that the set of stationary points is monotonically decreasing with the depth w.r.t the containment relation.

\begin{theorem} \label{th:monotoneStatPointSet}[Monotonocity of stationary points sets] The set of stationary policies of PGTS decreases with depth \ie
     \[\Pi^{m}\subseteq\Pi^{m-1}.\]
And set stationary policies of PGTS with infinite depth,  $\Pi^\infty := \bigcap_{m=0}^{\infty}\Pi^m = \argmax_{\pi}J^\pi$ contains only global optimal policies.     
\end{theorem}
The result shows that PGTS has fewer and fewer local minimas as the depth increase, as illustrated in Table \ref{tb:stationaryPolicy}. This supports the use of deeper PGTS in difficult MDPs.

\begin{algorithm}[H]
\caption{PGTS Algorithm}
\label{main:alg:PGTS}
\textbf{Input}: Depth $m$, stepsizes $\eta_k$, initial policy $\pi_0$  
\begin{algorithmic}[1] %[1] enables line numbers
\WHILE{not converged; $k = k+1$}
   \STATE  Estimate Q-values $Q^{\pi_k}$  
   \STATE Update $\pi_{k+1}(|s) = proj\Bigm[\pi_k(|s)+\eta_k d^{\pi_{k}}(s)(T^mQ^{\pi_k})(s,\cdot) \Bigm],\qquad \forall s  \quad s.t.\quad d^{\pi_k}(s)>0.$
\ENDWHILE
\end{algorithmic}
\end{algorithm}

\begin{table}[]
\centering
\begin{tabular}{c|c|c}
&  Ladder MDP & Tightrope MDP\\
    \toprule
$\Pi^0$ & $\Pi^1\cup \{\pi_{xyz00},\pi_{00xyz},\pi_{xy00z},\pi_{x00yz}\mid 0<x,y,z\leq1\}$ & $\{\pi_{00xy},\pi_{11xy}\mid 0\leq x,y\leq 1\}$\\\\
$\Pi^1$ & $\Pi^2\bigcup\Bigm\{\pi_{0x0yz},\pi_{xy0z0}\mid  0<x,y,z\leq1\Bigm\}$&$ \Bigm\{\pi_{11xy}\mid 0\leq x,y\leq 1\Bigm\}$\\\\
$\Pi^2$ & $\Pi^3\bigcup\Bigm\{\pi_{x0yz0},\pi_{ 0xy0z}\mid 0<x,y,z\leq1\Bigm\}$ &$\Pi^1$ \\\\
$\Pi^3$ & $\Pi^4\bigcup\Bigm\{\pi_{0xyz0}\mid  0\leq x,y,z\leq 1\Bigm\}$ &$\Pi^1$\\\\
$\Pi^4$ & $\Bigm\{\pi_{11111}\Bigm\}$ &$\Pi^1$\\
\bottomrule
\end{tabular}
    \caption{Set of Stationary Policies : Initial state $s_0$  and $\pi_{a_0a_1a_2a_3a_4}(\text{'right'}|s_i) = a_i$ for all $i$.}    \label{tb:stationaryPolicy}
\end{table}

A direct corollary of this theorem is that as we increase lookahead depth, the worst local minima gets better. Formally, we have:
\begin{corollary}\label{MonotonicWorstOptima}
Define the worst local minima to be $b^m = \min_{\pi \in \Pi^m}  J^\pi$, then we have: $
b^m \ge b^{m-1}$
\end{corollary}
We conclude that increasing the depth of PGTS improves the quality of solutions. However, several open questions remain for future work:

\begin{itemize}
    \item Does PGTS require infinite depth to eliminate all local optima, or is there a finite depth \( M \) such that \( \Pi^M = \Pi^\infty \)? Alternatively, can we identify a threshold \( M_\epsilon \) that guarantees \(\epsilon\)-suboptimality in the worst case, i.e., \(\min_{\pi \in \Pi^{M_\epsilon}} J^\pi \geq J^* - \epsilon\)? We hypothesize that \( M_\epsilon \) depends on structural properties of the MDP, such as diameter, connectivity, or reward sparsity.
    \item Does increasing the depth of PGTS accelerate convergence compared to standard PG? While experiments suggest that depth improves convergence, formal analysis is needed to confirm whether finite depth with appropriately small learning rates outperforms standard PG, particularly given prior findings that infinite-depth PGTS converges in \( |S| \) steps \cite{PGTS}.
\end{itemize}

\section{Conclusion}
We introduced Policy Gradient with Tree Search (PGTS), which incorporates lookahead into the policy gradient framework to address convergence to suboptimal local optima. Theoretical analysis shows that increasing tree search depth reduces the set of undesirable stationary points, improving the worst-case stationary policy. This guarantees eventual convergence to global optima with sufficient depth, even when policy updates are limited to visited states.

Empirical evaluations on Ladder, Tightrope, and Gridworld MDPs support these findings. PGTS demonstrates "farsighted" behavior, escaping local optima where standard PG fails. By navigating complex reward landscapes and tolerating temporary performance declines, PGTS achieves superior long-term solutions. While PGTS performs comparably to PG in well-connected MDPs, its advantages are pronounced in environments with large diameters or sparse rewards, where deeper lookahead facilitates faster reward propagation and optimal path discovery.

Future directions include developing sample-efficient PGTS algorithms for high-dimensional, model-free settings and establishing convergence rates with novel analytical tools. Investigating the relationship between lookahead depth and MDP structural properties—such as diameter, connectivity, and reward sparsity—can provide further insights for practical implementation.

In summary, PGTS offers a robust framework for overcoming local optima in policy gradient methods, enabling the discovery of higher-quality solutions in complex reinforcement learning tasks.

\bibliography{main}
\bibliographystyle{plain}

\newpage

\appendix
\section{Proofs}
\subsection{Proof of Lemma~\ref{rs:PiEquivalance}}
\begin{proof}
%     For $\pi \in \Pi^m_\infty$ if and only if
% \begin{align}
%     \pi(\cdot|s)=\lim_{\eta\to\infty}Proj\Bigm[\pi(\cdot|s) +\eta U^m_\pi(s,\cdot) \Bigm] \in \argmax_{ x\in \Delta_{\A} }\innorm{x,U^m_\pi(s,\cdot)}, \qquad \text{(from basic simplex projection)}.
%     \end{align}
First, we note that the claim in the lemma is trivial for $\eta=\infty$. \\
For any fixed $s\in\St, \pi \in \Pi^m_\eta$ ,  let $u :=\eta d^\pi(s) T^m Q^\pi(s,\cdot) $  and $f(x) :=
     \innorm{u,x}-\frac{1}{2}\norm{\pi_s-x}^2$. Then, we have $
 \pi_s=Proj\Bigm[\pi_s + u \Bigm]\in\argmax_{x}f(x).$ Let $x^* \in \argmax_{x}\innorm{u,x}$ and $\alpha\in (0,1]$, then
\begin{align*}
    \innorm{u,\pi_s} 
        &= f(\pi_s) 
        = \max_x f(x) 
        \geq f\left(\left(1-\alpha\right)\pi_s+\alpha x^*\right) \\
    &= \innorm{u,\pi_s} 
        - \alpha \innorm{u,\pi_s - x^*} 
        - \frac{\alpha^2}{2}\norm{\pi_s - x^* }^2,
        \qquad\text{(put values in $f$)} \\
    \implies 
    &\frac{\alpha}{2}\norm{\pi_s - x^* }^2 
        \geq \innorm{u,x^*-\pi_s} 
        = \max_x \innorm{u,x} - \innorm{u,\pi_s}.
\end{align*}
We now analyze the result by considering two possible cases for $\pi_s$.

\textbf{Case 1:} $\pi_s =x^*$ then we have $\pi_s \in \argmax_{x}\innorm{u,x}$.

\textbf{Case 2:} $\pi\neq x^*$, then the above relation implies $\alpha \geq 2\frac{ \max_x \innorm{u,x}-\innorm{u,\pi_s}}{\norm{\pi_s - x^* }^2}$ for all $\alpha \in (0,1]$. This implies $\max_x \innorm{u,x}=\innorm{u,\pi_s}$.

Both cases imply $\pi_s \in \argmax_{x}\innorm{u,x} = \argmax_{x}\innorm{ d^\pi(s) T^m Q^\pi(s,\cdot),x} $.
To conclude,  $\Pi^m_\eta =\{\pi\mid \pi_s \in \argmax_{x}d^\pi(s)\innorm{ T^mQ^\pi(s,\cdot),x},\quad \forall s\in\St\} $ for all $\eta$.
\end{proof}
\subsection{Proof of Proposition~\ref{prop:allDepthContainOptimal}}
We note that while this proposition is a corollary of Theorem ~\ref{th:monotoneStatPointSet}, for completeness we include here a simpler proof that does not depend on the theorem.
\begin{proof}
    Let $\pi^*$ an optimal policy. From the optimality conditions of $\pi^*$ and the optimal Q-function $Q^*$ we get for all $s\in \sigma_{\pi^*}$:
    \begin{align*}
        \pi^*_s &\in \argmax_{\pi'_s} \innorm{\pi'_s, Q^*(s,\cdot)}\\& = \argmax_{\pi'_s} \innorm{\pi'_s, T^mQ^*(s,\cdot)}
    \end{align*}
    and from Lemma~\ref{rs:PiEquivalance} we deduce that $\pi^*\in \Pi^m$.
\end{proof}
\subsection{Proof of Lemma~\ref{rs:optimalityCondn}}
\begin{proof}  
We begin by showing that if $\pi \in \Pi^m$, then the equality described in the lemma holds.
Fix some $\pi \in \Pi^m$. We first prove that $\max_{a}(T^mQ^\pi)(s,a) \leq v^\pi(s)$ for all $s\in\sigma_\pi$. Let $s\in\sigma_\pi$, and observe that:
\begin{align*}
    \omega(s) :&=\max_{a}(T^mQ^\pi)(s,a)\\&=\innorm{\pi_s,(T^mQ^\pi)(s,\cdot)},\qquad \text{(from Lemma \ref{rs:PiEquivalance}, as $s\in\sigma_\pi$)}\\&=\innorm{\pi_s,(T^mT^\pi Q^\pi)(s,\cdot)},\qquad \text{(as $T^\pi Q^\pi = Q^\pi$)}
    \\&\leq \innorm{\pi_s,(T^{m+1} Q^\pi)(s,\cdot)},\qquad \text{(as $T Q^\pi = \max _{\pi'}T^{\pi'} Q^\pi \succeq T^\pi Q^\pi $, and   $Tv\geq Tu$ if $v\succeq u$)}\\
    &= \sum_{a}\pi(a|s)\Bigm[(T^{m+1} Q^\pi)(s,a)\Bigm],\qquad\text{( expanding the dot product)}\\&= \sum_{a}\pi(a|s)\Bigm[R(s,a)+ \gamma \sum_{s'}P(s'|s,a)\max_{a'}(T^{m} Q^\pi))(s',a')\Bigm],\qquad \text{(defn of $T$)}\\
    &= \sum_{a}\pi(a|s)\Bigm[R(s,a)+ \gamma  \sum_{s'}P(s'|s,a)\omega(s')\Bigm],\qquad\text{(defn. of $\omega$ )}\\&= R^\pi(s)+ \gamma P^\pi(\cdot|s)\omega,\qquad \text{(compactifying )}.
\end{align*}
Note that the support set $\sigma_\pi$ is closed under $P^\pi$, that is,  $s\in\sigma_\pi, P^\pi(s'|s)>0$ implies $s'\in\sigma_\pi$. Hence we get
\begin{align}
\omega\restriction_{\sigma_\pi} &
\leq R^\pi\restriction_{\sigma_\pi}+ \gamma P^\pi\restriction_{\sigma_\pi}\omega\restriction_{\sigma_\pi},\qquad \text{(where $x\restriction_{\sigma_\pi}\in\R^{\sigma_\pi}$ denotes $x$ restricted to the set $\sigma_\pi$)}.
\end{align}
Observe that $P^\pi\restriction_{\sigma_\pi}\in \R^{\sigma_\pi\times\sigma_\pi}$ is a valid stochastic matrix. So, by unrolling the  the above equation infinite times, we get 
\begin{align}
\omega\restriction_{\sigma_\pi} & \leq\sum_{n=0}^{\infty}\gamma^n (P^\pi\restriction_{\sigma_\pi})^nR^\pi\restriction_{\sigma_\pi} = v^\pi\restriction_{\sigma_\pi}.
\end{align}
We have so for proved $\max_{a}(T^mQ^\pi)(s,a) \leq v^\pi(s)$ for all $s\in\sigma_\pi$. We proceed to prove the inverse inequality:

\begin{align}
    v^\pi (s) &= \innorm{\pi_s, Q^\pi(s,\cdot)},\qquad\text{(basic RL)}\\&= \innorm{\pi_s,T^\pi Q^\pi)(s,\cdot)},\qquad\text{ as $T^\pi Q^\pi = Q^\pi$}\\&\leq \max_{a}(T^mQ^\pi)(s,a),\qquad \text{(as $TQ = \max _{\pi }T^\pi Q \succeq Q$).}
\end{align}

Hence, we conclude $v^\pi (s) = \max_{a}(T^mQ^\pi)(s,a)$ for all $s\in\sigma_\pi$. 

We now continue to prove the converse direction of the theorem. Assume that for all $s\in\sigma_\pi$, we have
\begin{align}
  v^\pi(s) &= \max_{\pi'_s}  \innorm{\pi'_s,(T^mQ^\pi)(s,\cdot)}\\&=\max_{\pi_0}\max_{\pi_i}\innorm{\pi_0(\cdot|s),(\bigcirc_{i=1}^{m}T^{\pi_i}Q^\pi)(s,\cdot)},\qquad \text{(as $T Q = \max_{\pi} T^\pi Q$)}.
\end{align}
Where $\bigcirc_{i=1}^m$ is used to denote sequential composition. We know
 $v^\pi(s) = \innorm{\pi(\cdot|s),((T^{\pi})^m Q^\pi)(s,\cdot)}$, that is, the  expression  $\innorm{\pi_0(\cdot|s),(\bigcirc_{i=1}^{m}T^{\pi_i}Q^\pi)(s,\cdot)}$ is maximized for $\pi_i = \pi$. In particular, this implies, 
 \begin{align}
    \pi_s &\in \argmax_{(\pi_0)_s}\max_{\pi_i}\innorm{\pi_0(\cdot|s),(\bigcirc_{i=1}^{m}T^{\pi_i}Q^\pi)(s,\cdot)}\\&=\argmax_{(\pi_0)_s}\innorm{\pi_0(\cdot|s),(T^{m}Q^\pi)(s,\cdot)},\qquad \text{(as $T Q = \max_{\pi} T^\pi Q$)}.
 \end{align}
 This implies, $\pi\in\Pi^m$.
\end{proof}

\subsection{Proof of Theorem~\ref{th:monotoneStatPointSet}} 
\begin{proof}  
Let $\pi \in \Pi^m$, then from Lemma \ref{rs:PiEquivalance}, we have 
\[\pi_s  
\in\argmax_{\pi'_s}\innorm{\pi'_s,(T^mQ^\pi)(s,\cdot)},\qquad \forall s\in \sigma_\pi.
\]
And from optimality condition in Lemma \ref{rs:optimalityCondn} (as $s\in\sigma_\pi$) we have
\begin{align*}
 v^\pi(s) &=\max_{\pi'_s}\innorm{\pi'_s,(T^mQ^\pi)(s,\cdot)}\\&=\max_{\pi_0}\max_{\pi_i}\innorm{\pi_0(\cdot|s),(\bigcirc_{i=1}^{m}T^{\pi_i}Q^\pi)(s,\cdot)},\qquad \text{(as $T Q = \max_{\pi} T^\pi Q$)}\\
&\ge \max_{\pi_0}\max_{\pi_i}\innorm{\pi_0(\cdot|s),(\bigcirc_{i=1}^{m-1}T^{\pi_i}T^\pi Q^\pi)(s,\cdot)} \\&=\max_{\pi_0}\max_{\pi_i}\innorm{\pi_0(\cdot|s),(\bigcirc_{i=1}^{m-1}T^{\pi_i} Q^\pi)(s,\cdot)},\qquad \text{(as $T^\pi Q^\pi= Q^\pi$)}\\
&=\max_{\pi'_s}\innorm{\pi'_s,(T^{m-1} Q^\pi)(s,\cdot)}\\&=\max_{a}(T^{m-1} Q^\pi)(s,a)\\& = T^m v^\pi (s)\\&\ge v^\pi(s), \qquad \text{(as $\max_{\pi}T^\pi Q^\pi= T Q^\pi$ is monotone)}
\end{align*} The above specifically implies $v^\pi(s) =\max_{a}(T^{m-1} Q^\pi)(s,a)$, therefore from Lemma \ref{rs:optimalityCondn} we have $\pi\in\Pi^{m-1}$. Finally, from theorem 1 in \cite{kumar2024policy} we know that for $m,\eta=\infty$ we converge to a globally optimal policy, this along with Lemma \ref{rs:PiEquivalance} implies that $\Pi^\infty$ contains only optimal policies.
\end{proof}
\end{document}